\pdfoutput=1

\documentclass[11pt]{article}

\usepackage[final]{acl}

\usepackage{times}
\usepackage{latexsym}
\usepackage[most]{tcolorbox}
\usetikzlibrary{shapes}
\usepackage[T1]{fontenc}

\usepackage[utf8]{inputenc}

\usepackage{microtype}

\usepackage{inconsolata}
\usepackage{booktabs}
\usepackage{multirow}
\usepackage{graphicx}
\usepackage{hyperref}
\usepackage{listings}
\usepackage[most]{tcolorbox}
\usepackage{xcolor}

\definecolor{myred}{HTML}{EC6460}   
\definecolor{myyellow}{HTML}{ECBD60  } 
\definecolor{mygreen}{HTML}{90EC60}  
\definecolor{mygray}{HTML}{AEB0AC}   
\definecolor{deeppurple}{HTML}{3c3273}
\definecolor{mygray}{RGB}{240,240,240}
\definecolor{darkgray}{RGB}{64,64,64} 

\newtcolorbox{mycodebox}{
    enhanced,
    fonttitle=\bfseries,
    colback=white,
    colframe=mygray,
    colbacktitle=mygray,
    coltitle=black,
    rounded corners,
    boxrule=0.5mm, 
    top=5mm, 
    interior style={top color=mygray,bottom color=white}, 
    overlay={
        \node[anchor=west, font=\bfseries] at ([xshift=5pt, yshift=-10pt]frame.north west) 
        {\textcolor{myred}{$\bullet$} \textcolor{myyellow}{$\bullet$} \textcolor{mygreen}{$\bullet$}};
    },
    left=4mm,
    right=2mm,
    bottom=2mm,
}

\newtcolorbox{myshellbox}{
    enhanced,
    fonttitle=\bfseries,
    colback=white,
    colframe=mygray,
    coltext=white,
    colbacktitle=mygray,
    coltitle=black,
    rounded corners,
    boxrule=0.5mm, 
    top=5mm, 
    interior style={top color=black,bottom color=darkgray}, 
    overlay={
        \node[anchor=west, font=\bfseries] at ([xshift=5pt, yshift=-10pt]frame.north west) 
        {\textcolor{myred}{$\bullet$} \textcolor{myyellow}{$\bullet$} \textcolor{mygreen}{$\bullet$}};
    },
    left=2mm,
    right=2mm,
    bottom=2mm,
}

\title{When LLMs are Unfit Use \fastfit{}:\\ Fast and Effective Text Classification with Many Classes}

\author{Asaf Yehudai$^1{}^2$ 
    \qquad Elron Bandel$^1$\\
$^1$IBM Research, $^2$Hebrew University of Jerusalem\\
\texttt{\{first.last\}@ibm.com}\\
}
\newcommand{\fastfit}{\textit{FastFit}}
\newcommand{\fewmany}{\textit{FewMany}}

\begin{document}
\maketitle
\begin{abstract}
We present \fastfit{}, a method, and a Python package design to provide fast and accurate few-shot classification, especially for scenarios with many semantically similar classes. \fastfit{} utilizes a novel approach integrating batch contrastive learning and token-level similarity score.  Compared to existing few-shot learning packages, such as SetFit, Transformers, or few-shot prompting of large language models via API calls, \fastfit{} significantly improves multi-class classification performance in speed and accuracy across \fewmany{}, our newly curated English benchmark, and Multilingual datasets. \fastfit{} demonstrates a 3-20x improvement in training speed, completing training in just a few seconds. The \fastfit{} package is now available on GitHub and PyPi, presenting a user-friendly solution for NLP practitioners.\\
Code: \url{https://github.com/IBM/fastfit}
Data: \url{https://huggingface.co/FastFit}
\end{abstract}

\section{Introduction}
Few-shot classification presents a unique challenge, especially when dealing with a multitude of classes that share similar semantic meanings. Expanding the training data can be both time-consuming and costly. To address this challenge, two primary categories of tools have been developed: few-shot prompting of large language models (LLMs) via API calls, or packages designed for fine-tuning smaller language models using the limited available data. However, we recognize the drawbacks of applying both approaches in practice.

Few-shot prompting of LLMs leverages their multitasking abilities to tackle data scarcity. However, in the presence of many classes, LLMs encounter three major challenges: (1) LLMs struggle to incorporate demonstrations of all classes within their context window. (2) Utilization of the long context for the classification task can be challenging \cite{liu2023lost}. (3) Inference time is slow due to model size, and prompt length.

In contrast, the approach of fine-tuning smaller language models capitalizes on their adaptability to specific tasks, as demonstrated to be effective in recent works. However, these methods can be challenging to deploy as they require architectural adjustments \cite{yehudai2023qaid} or, like SetFit, may prove less suitable for classification with many classes \cite{tunstall2022efficient}.

\begin{figure}
    \centering
    \resizebox{\columnwidth}{!}{%
        \includegraphics{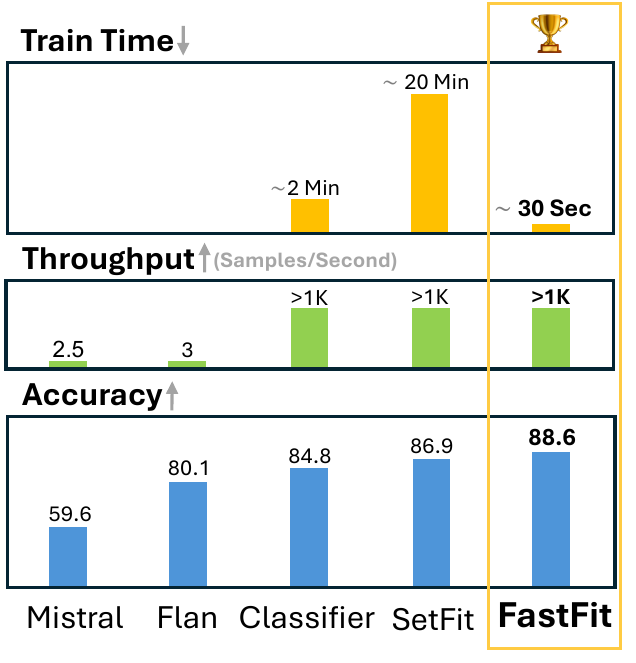}
    }
    \caption{\fastfit{} achieves SOTA classification results combined with fast training and high throughput. Outpreforming other fine-tuning methods and strong LLMs.}
    \label{fig:fastfit_summary}\vspace{-1em}
\end{figure}

In this work, we present \fastfit{}, a fast and accurate method, and a pip-installable Python package designed for fine-tuning small language models in few-shot classification tasks involving many classes. Through various experiments, on our newly curated \fewmany{} benchmark, we demonstrate that \fastfit{} training is significantly faster, providing a 3-20x speedup. This enables training within seconds, as illustrated in Fig. \ref{fig:fastfit_summary}. \fastfit{} outperforms earlier packages, including SetFit, Transformer, and multi-task models like FLAN, or larger LLMs like LLama-70B, in both English and Multilingual settings.

The core contribution facilitating this speedup and improvement lies in \fastfit{}'s use of batch contrastive training, recognized for its efficiency and effectiveness \cite{khosla2021supervised}. This technique brings same-class texts closer while pushing apart all other texts. \fastfit{} also incorporates token-level text similarity measures that leverage fine-grained information \cite{zhang2020bertscore, khattab2020colbert}. Additionally, we integrate text augmentation techniques to enhance the robustness of the training process \cite{gao-etal-2021-simcse}.

The \fastfit{} package is easy to install and use, interfacing with standard training APIs (See \S\ref{fastfit_api}). We hope that \fastfit{} will help make text classification easier and faster for the benefit of the whole community.

\section{The \fastfit{} Library}
\label{fastfit_api}
The \fastfit{} Python package is available on PyPI and can be installed with: 
\begin{myshellbox}
\begin{lstlisting}
 $ pip install fast-fit
\end{lstlisting}
\end{myshellbox}





To utilize \fastfit{}, import the \fastfit{} trainer, which inherits from the Hugging Face (HF) trainer. This enables \fastfit{} to be customizable, inheriting all parameters from the HF trainer.
\fastfit{} supports loading datasets either by directly passing the dataset or providing file paths.

Here is a simple code example of loading and training \fastfit{}. In App. \S\ref{sec:app:code_exmp}, we provide a complete code example.







    






\begin{mycodebox}
\begin{lstlisting}
from fastfit import FastFitTrainer

trainer = FastFitTrainer(
    model_name_or_path=
      "roberta-large",
    text_column_name="text",
    label_column_name="label",
    dataset=dataset,
)

model = trainer.train()
results = trainer.evaluate()
\end{lstlisting} 
\end{mycodebox}

 As \fastfit{} utilizes example texts and class names, it expects the data to have \texttt{text} and \texttt{label} fields or to map the existing fields to them using the \texttt{label\_column\_name} and \texttt{text\_column\_name} parameters of the \texttt{FastFitTrainer}. Our trainer also supports training with either CLS or token-level similarity metrics, set by the \texttt{sim\_rep} parameter. The trainer allows to modify the number of augmentation repetitions with the \texttt{num\_repeats} parameter.
Then after training, we can easily save the model:

\begin{mycodebox}
\begin{lstlisting}
model.save_pretrained("fast-fit")
\end{lstlisting}
\end{mycodebox}

And later load it for inference, See App. \S\ref{sec:app:code_exmp}.

\section{Method}

Given a few-shot text classification dataset containing texts and their corresponding classes denoted as $\{{x_i, y_i}\}_{i=1}^N$, let $C = \{c_j\}_{j=1}^M$ represent all possible classes. Our task is to classify each $x_i$ into a class $y_i \in C$. To achieve this goal we aim to encode both texts and class names into a shared embedding space, where they are represented closely, according to a similarity metric $S$, when they belong to the same class and are represented further apart when they do not. To accomplish this, we optimize the following batch contrastive loss:


\begin{equation}\label{eq:sup_loss}
    \mathcal{L} = \sum_{b\in [B]} \frac{-1}{|P(b)|}  
    \sum_{p\in P(b)} \log \frac{e^{S(x^{b}, x^{p})/\tau}}{\sum_{a \in [B]\setminus b} e^{S(x^b, x^{a})/\tau}}
\end{equation}

Here, $\{x_b\}_{b=1}^B$ represents a batch of $B$ texts, and $P(b)$ refers to the set of texts in the same class as $b$ in the batch, given by $P(b) = \{c \in [B] ,|, y_c=y_b\}$. The function $S$ is the similarity metric, and $\tau$ is a scalar temperature parameter regulating the penalty for negative texts.

For each text in the batch, we augment the batch by including its class name as an additional example. Additionally, we repeat the texts in the batch $r$ times as a data augmentation technique, following \citet{gao-etal-2021-simcse} by treating the dropout as a minimal augmentation at the representation level. This method has demonstrated significant success in generating sentence embeddings, and we leverage it here to enhance representation for text classification.


In our data-scarce setting, we employ fine-grained token-level similarity metrics, leveraging textual details. This approach, successful in works like BERT-Score and ColBERT, defines the similarity metric between texts $x_i$ and $x_j$ as the sum of cosine similarities between $x_i$ and the most similar tokens in $x_j$. Specifically, with tokens denoted as $x_i^1, \ldots, x_i^n$ and $x_j^1, \ldots, x_j^m$ respectively, the similarity score is computed as follows:

\begin{equation}\label{eq:tok_sim}
S(x_i, x_j) = \sum_{k=1}^{n} \max_{l=1}^{m} E_{\theta}(x_i^k) \cdot E_{\theta}(x_j^l)
\end{equation}
where $E_\theta(x_i^k)$ is a dense representation of token $x_i^k$ produced by a parametric encoder model with parameters $\theta$.

During inference, when provided with a new text, $x_u$ we classify it to the most similar class $y_i \in C$ with respect to a similarity metric $S$. This method draws inspiration from the way inference is conducted in retrieval systems, eliminating the need for a classification head and aligning the training and inference objectives.

\section{\fewmany{} Benchmark}
To rigorously evaluate the capabilities of models in few-shot text classification with many classes, we introduce the \fewmany{} benchmark, a collection of eight diverse classification datasets, each featuring at least 50 classes.
The benchmark spans several domains, including intent detection, topic classification, question classification, and product classification. Each domain in \fewmany{} presents a unique input type, such as short informal user queries, arguments, claims, long-form Wikipedia articles, questions, and product descriptions. By covering a wide spectrum of cases, \fewmany{} enables a comprehensive evaluation of model performance in distinguishing between many semantically similar classes, often with subtle distinctions. In this work, we conduct experiments on \fewmany{} under 5-shot and 10-shot scenarios, where the $k$-shot scenario refers to a training set with $k$ examples per class. Further details and data statistics can be found in Appendix \ref{sec:app:data_stat}.

\section{Experiments}

\begin{table*}[t!]
\centering
\begin{tabular}{@{}lcccccccccr@{}}
\toprule
\multicolumn{1}{c}{Method} & Model & C150 & AP106 & B77 & AT71 & DB70 & HU64 & CS55 & T50 & Average \\ \midrule
& & \multicolumn{8}{c}{5-shot} & \\ \midrule
\multirow{2}{*}{\fastfit{}} & S & 91.3 & 47.5 & 81.0 & 95.4 & 82.5 & 82.2 & 86.1 & 80.3 & 80.8 \\
& L & \textbf{93.4*} & \textbf{50.9*} & \textbf{85.2*} & \textbf{96.2} & \textbf{83.1*} & \textbf{84.6*} & 87.5 & \textbf{84.8*} & \textbf{83.2*} \\
\midrule
\multirow{2}{*}{SetFit} & S & 89.0 & 45.9 & 77.3 & 94.8 & 79.0 & 80.0 & 84.1 & 79.5 & 78.7 \\
& L & 90.4 & 48.2 & 81.7 & 95.6 & 80.1 & 81.9 & 87.8 & 83.9 & 81.2 \\
\midrule
\multirow{2}{*}{Classifier} & S & 86.3 & 30.4 & 68.2 & 95.1 & 70.5 & 73.9 & 82.6 & 63.4 & 71.3 \\
& L & 92.0 & 44.5 & 79.7 & 96.0 & 76.8 & 79.4 & \textbf{88.2} & 73.3 & 78.7 \\
\midrule
& & \multicolumn{8}{c}{10-shot} & \\ \midrule
\multirow{2}{*}{\fastfit{}} & S & 93.5 & 54.5 & 86.4 & 95.9 & 87.8 & 85.8 & 88.5 & 84.1 & 84.6 \\
& L & \textbf{95.3*} & \textbf{57.5} & \textbf{88.8*} & 96.5 & \textbf{88.7*} & \textbf{87.9*} & 89.4 & \textbf{88.0*} & \textbf{86.5*} \\
\midrule
\multirow{2}{*}{SetFit} & S & 90.9 & 53.6 & 84.8 & 95.5 & 85.9 & 85.1 & 87.7 & 83.7 & 83.4 \\
& L & 88.4 & 53.6 & 86.4 & 95.7 & 85.8 & 85.4 & 88.8 & 86.4 & 83.8 \\
\midrule
\multirow{2}{*}{Classifier} & S & 91.5 & 46.9 & 80.2 & 95.5 & 82.1 & 83.1 & 86.5 & 78.0 & 80.5 \\
& L & 94.5 & 57.1 & 87.4 & \textbf{96.6} & 87.0 & 86.0 & \textbf{90.9} & 86.8 & 85.8 \\
\bottomrule
\end{tabular}
\caption{Accuracy results of \fastfit{} and baselines on 5/10-shot text classification. Results show that \fastfit{} outperforms SetFit and a standard classifier. Moreover, \fastfit{} small is comparable to SetFit large. Results with * are statistically significant by t-test (p < 0.05) compared to the large standard classifier.}
\label{tbl:fewmany}
\end{table*}

\begin{table*}[t!]
\centering
\begin{tabular}{@{}llllllll@{}}
\toprule
Model & C150 & B77 & AT71 & HU64 & CS55 & T50 & Avg. \\ \midrule
Flan-ul2 & 80.3 & 71.5 & \textbf{97.3} & \textbf{76.2} & \textbf{89.4} & 65.6 & \textbf{80.1} \\
Flan-XXL & \textbf{82.1} & \textbf{72.1} & 97.0 & 74.9 & 49.0 & \textbf{84.9} & 76.7 \\
Llama-2-13B-chat & 53.0 & 42.6 & 77.0 & 53.2 & 54.8 & 49.6 & 55.0 \\
Llama-2-70B-chat & 60.8 & 45.7 & 88.9 & 62.8 & 57.9 & 37.7 & 59.0 \\
Mistral-7B & 63.5 & 46.8 & 87.0 & 71.7 & 58.8 & 29.5 & 59.6 \\ \midrule
\fastfit{} & \textbf{93.4} & \textbf{85.2} & \textbf{96.2} & \textbf{84.6} & 87.5 & \textbf{84.8} & \textbf{88.6} \\
SetFit & 90.4 & 81.7 & 95.6 & 81.9 & 87.8 & 83.9 & 86.9 \\
Classifier & 92.0 & 79.7 & 96.0 & 79.4 & \textbf{88.2} & 73.3 & 84.8 \\
\bottomrule
\end{tabular}
\caption{Accuracy results of a few LLMs models on 6 test sets from \fewmany{}. The Flan models outperform the other LLMs on datasets from \fewmany{}. Llama-70B scores higher than Llama-13B and is comparable to Mistral on these datasets. For comparison, we present the 5-shot results of the fine-tuning methods. }
\label{tbl:llm-result}
\end{table*}

\subsection{Baselines}\label{exp-eng:base}
We compare \fastfit{} with a few classification methods, including fine-tuning methods, like Standard and SetFit classifiers, and few-shot promoting of LLMs including Flan-XXL \cite{wei2022finetuned}, Flan-ul2 \cite{tay2023ul2}, llama-2-70b-chat \cite{touvron2023llama}, and Mistral-7b \cite{jiang2023mistral}. For all fine-tuning methods, we use small and large versions, where small is MPNet (110M parameters) \cite{song2020mpnet}, and large is Roberta-large (355M parameters) \cite{liu2019roberta} or equivalent.

\textbf{Standard Classifier.} A simple yet strong baseline is a standard fine-tuning of an encoder-only model.  Since we assume no validation sets, we use best practices as described in previous works, and train for 40 epochs, with a learning rate of $1e-5$, and batch size of 16 \cite{Lin2023SelectiveID}. We recovered runs that didn't converge.
 
\textbf{SetFit.} Sentence Transformer Fine-tuning (SetFit) \cite{tunstall2022efficient} is a two-stage method for training a Sentence Transformer model \cite{reimers2019sentencebert}, specifically designed for few-shot classification tasks. In the first stage, the encoder undergoes fine-tuning using triplet loss, and in the second stage, the classification head is trained. For the small model we use \texttt{paraphrase-mpnet-base-v2}\footnote{\href{https://huggingface.co/sentence-transformers/paraphrase-mpnet-base-v2}{ST-MPNet}}, and for the large model, we used \texttt{all-Roberta-Large-v1}\footnote{\href{https://huggingface.co/sentence-transformers/all-roberta-large-v1}{ST-Roberta-Large}}, both trained with sentence transformer objective before. We trained the model with a learning rate of $1e-5$, a batch size of 16, for one epoch, based on the parameters defined in SetFit's paper.

\textbf{Flan.} Flan language models are fine-tuned on a diverse range of NLP tasks and datasets, making them adaptable for various NLP tasks in a few-shot manner. Here, we experimented with Flan-XXL (11B) and Flan-ul2 (20B) models. These models have a 4K tokens context window.


\textbf{Llama.} Llama-2-chat is a set of large language models developed for conversational applications and has strong multi-task few-shot capabilities. Here, we experimented with a Llama model that supports a 4K tokens context window.

\textbf{Mistral.} Mistral is a strong 7B open-source large language model. Here, we used the instruct-tuned version. Mistral supports an 8K tokens context window.

\subsection{Experimental Setup}\label{exp-eng:setup}
\textbf{Training Setup.} We fine-tune the \fastfit{} model with a learning rate of $1e-5$, a batch size of 32, and a maximum sequence length of 128 tokens, for 40 epochs. We used AdamW optimizer, 16-bit floating-point (FP16) precision, and applied 4 batch repetitions that act as augmentations.

For all LLMs, we fit the maximum possible number of examples into their context window. For 
AP106 and DB70 test sets even a 1-shot example do not fit into the context. Hence we compare LLM results on the remaining six test sets.

\textbf{Evaluation Setup.} Few-shot evaluations can be noisy due to variations in the small datasets \cite{dodge2020finetuning, zhang2021revisiting}. To address this challenge, we perform all our experiments using 5 random training split variations and report the mean results. 

\begin{table*}[t!]
\centering

\begin{tabular}{@{}p{2cm}p{1.5cm}p{1.2cm}p{1.2cm}p{1.2cm}p{1.2cm}p{1.2cm}p{1.2cm}p{1.5cm}@{}}
\toprule
\multicolumn{1}{l}{Method} & Size  & En   & De   & Ja   & Es   & Fr   & Zh   & Average \\ \midrule
                           &       & \multicolumn{6}{c}{5-shot}              &         \\ \midrule
\multirow{2}{*}{\fastfit{}}& S & \underline{72.3} & \underline{65.0} & \underline{68.7} & \underline{65.9} & \underline{68.0} & \underline{68.4} & \underline{68.1} \\
                        & L & \textbf{77.6*} & \textbf{70.5*} & \textbf{73.7*} & \textbf{71.7*} & \textbf{73.1*} & \textbf{73.7*} & \textbf{73.4*}  \\
SetFit & S & 67.9 & 62.2 & 66.8 & 64.0 & 65.0 & 66.7 & 65.4   \\
\multirow{2}{*}{Classifier}& S & 61.2 & 56.8 & 59.7 & 58.4 & 59.8 & 61.4 & 59.5   \\
                        & L & 66.4 & 56.0 & 65.3 & 56.6 & 60.0 & 61.9 & 61.0      \\ \midrule
                        &       & \multicolumn{6}{c}{10-shot}             &         \\ \midrule
\multirow{2}{*}{\fastfit{}}& S & \underline{77.6} & 70.5 & 73.7 & \underline{71.7} & \underline{73.1} & \underline{73.7} & \underline{73.4}    \\
                        & L & \textbf{79.2*} & \textbf{74.8*} & \textbf{77.4} & \textbf{74.1*} & \textbf{75.7*} & \textbf{74.9*} & \textbf{76.0*}    \\
SetFit & S & 74.7 & 69.8 & 73.5 & 71.4 & 72.0 & 72.9 & 72.4    \\
\multirow{2}{*}{Classifier}& S & 72.2 & 67.7 & 71.0 & 68.6 & 69.7 & 70.0 & 69.9   \\
                        & L & 77.5 & \underline{71.2} & \underline{74.3} & 71.3 & 72.5 & 72.7 & 73.3     \\ \bottomrule
\end{tabular}
\caption{Accuracy results for \fastfit{} and baselines across six languages, under 5/10-shot settings. Results show that \fastfit{} consistently outperforms SetFit and the standard classifier. Notably, \fastfit{} small consistently surpasses SetFit's small and standard large classifiers. Results marked with an asterisk (*) are statistically significant according to t-test (p < 0.05) when compared to the large standard classifier.
}\vspace{-1em}
\label{tbl:multi-result}
\end{table*}

\subsection{Results}
In Table \ref{tbl:fewmany}, we present the results of \fastfit{}, SetFit, and the standard classifier for \fewmany{} eight datasets under 5/10-shot settings. \fastfit{} large outperforms SetFit by $2\%$ and the standard classifier by $4.5\%$, in the 5-shot case. In the 10-shot case, \fastfit{} outperforms SetFit, and a standard classifier by $2.7\%$ and $0.7\%$, respectively. Moreover, \fastfit{} small is comparable to SetFit large in 5-shot and outperforms it in 10-shot.
Notably, \fastfit{} shows greater improvement in the 5-shot case compared to the 10-shot case and for the small model compared to the large one.


Table \ref{tbl:llm-result} displays the results of few-shot prompting for several LLMs. The Flan models exhibit higher performance than other LLMs, likely due to the presence of many classification datasets in the Flan dataset\footnote{To the best of our knowledge, the Flan dataset includes only T50 from our test sets}. This observation aligns with findings in zero-shot classification \cite{gretz-etal-2023-zero}. Llama-70B outperforms Llama-13B, and is comparable to Mistral-7B's performance, possibly due to Mistral's larger context length, allowing it to incorporate more examples per class.


The results suggest that in our setting, where numerous classes are present, even the best-performing LLMs we tested (Flan's) underperform compared to large standard classifiers and face challenges compared to \fastfit{}. It's important to note that, due to the model's size and the length of the few-shot prompt, inference time can be slow, with throughput exceeding 1 second per input, in contrast to about 1 millisecond with \fastfit{}.


\section{Multilingual Experiments}
\subsection{Datasets}
\label{sec:multi-data}
To evaluate \fastfit{}'s multilingual classification abilities we adopt Amazon Multilingual MASSIVE dataset \cite{fitzgerald2022massive}. From the 51 available languages, we selected six typologically diverse languages: English, Japanese, German, French, Spanish, and Chinese. MASSIVE is a parallel dataset, with 60 classes (See App. \S\ref{sec:app:data_stat}). 

\subsection{Baselines}

 For multilingual training, we utilized paraphrase-multilingual-mpnet-base-v2 as a small model and XLM-Roberta-Large as a large model. Both models underwent pretraining in a large number of languages. Notably, to the best of our knowledge, there is no multilingual sentence transformer model equivalent to Roberta-Large for SetFit training. Monolingual and XLM-Roberta-Large models were tested, but they yielded lower performance than the small model; hence, their results are detailed in Appendix \S\ref{sec:app:multi_results}. In English experiments, we maintained the use of monolingual models (see \S\ref{exp-eng:base}), conducting training and evaluation with the same setup outlined in \S\ref{exp-eng:setup}.
 



\subsection{Results}
In Table \ref{tbl:multi-result}, we present the results on MASSIVE in 5/10-shot scenarios using \fastfit{}, SetFit, and the standard classifier. \fastfit{} consistently outperforms both SetFit and the standard classifier in both 5-shot and 10-shot settings, across small and large models. In the 5-shot scenario, \fastfit{} large achieves an $8\%$ improvement over SetFit small and a $12.4\%$ improvement over the standard classifier. Meanwhile, \fastfit{} small shows a $2.7\%$ improvement over SetFit small and a $7.1\%$ improvement over the standard classifier. In the 10-shot case, \fastfit{} large outperforms SetFit small by $3.6\%$ and the standard large classifier by $2.7\%$. Similarly, \fastfit{} small exhibits improvements of $1.9\%$ and $3.5\%$ over SetFit small and the standard classifier, respectively.

It is noteworthy that \fastfit{} demonstrates improvement when scaling from a small to a large model, with gains of 5.3\% and 2.6\% in the 5-shot and 10-shot settings, respectively. This enhancement highlights the fact that \fastfit{} is not model-specific and thus is highly flexible for different sizes and types of models, unlike SetFit. Such flexibility is particularly crucial in few-shot settings where limited examples are available, highlighting the potential to train enhanced classifiers using domain- or language-specific models. Moreover, if unlabeled or pairwise data is available, using it for pretraining can lead to even further improvement.

\section{Fast Training}

\begin{figure}
    \centering
    \resizebox{\columnwidth}{!}{%
        \includegraphics{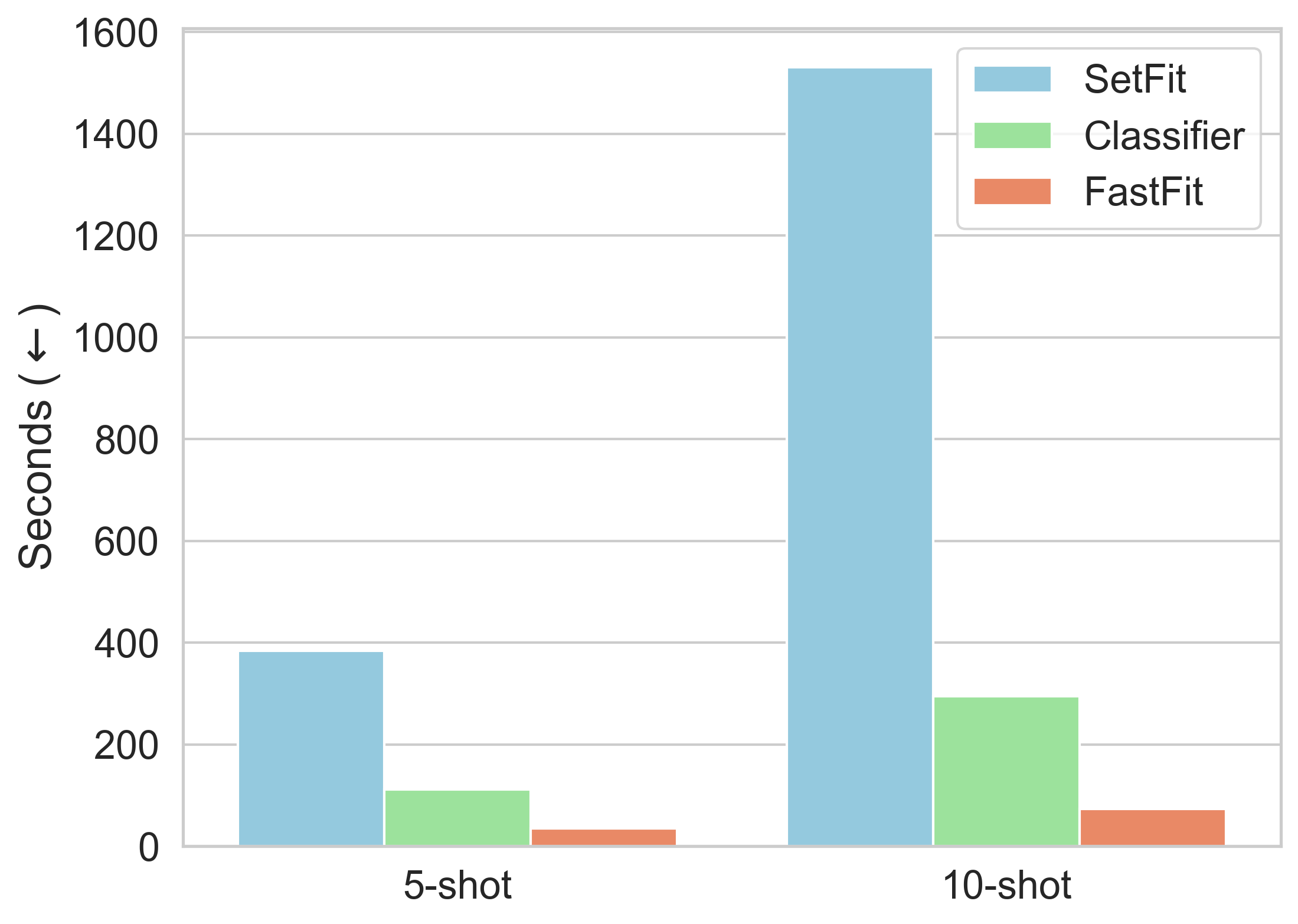}
    }
    \caption{Training times (sec) for \fastfit{}, SetFit, and standard classifier with MPNet model. \fastfit{} training is 3-20x faster.}
    \label{fig:fastfit}\vspace{-1em}
\end{figure}

\textbf{Training Times} for \fastfit{}, SetFit, and the standard classifier are illustrated in Figure \ref{fig:fastfit}. Results are average across all languages in MASSIVE, and all 5 seeds. \fastfit{} exhibits faster training times compared to both SetFit and the standard classifier, with a 3-20x decrease, and a training time of about 30 seconds (See more results at App. \S\ref{sec:app:runtime}). This can be attributed to a combination of technical and methodological factors. The improved implementation includes pre-training tokenization and FP16 training. Furthermore, the methodological advantage stems from using batch contrastive training, which leverages in-batch examples as negatives, in contrast to the triplet loss utilized by SetFit.

\begin{figure}[ht]
    \centering
    \resizebox{1.05\columnwidth}{!}{%
        \includegraphics{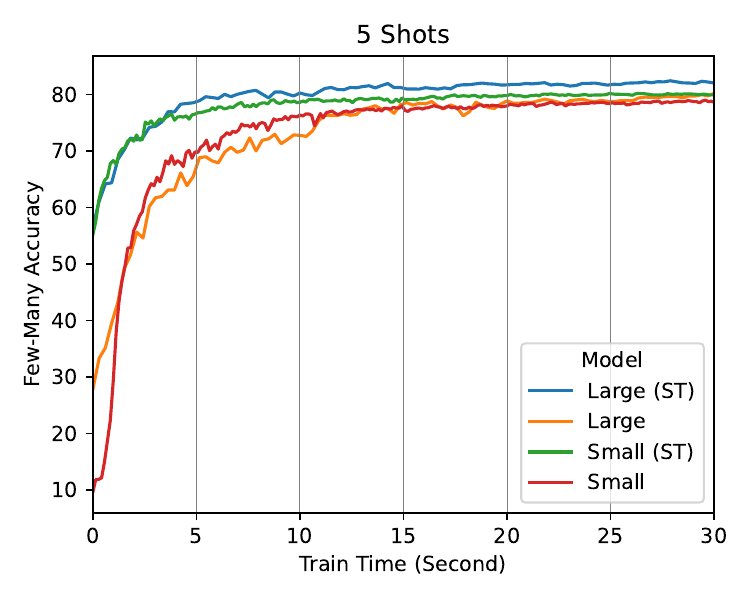}%
    }
    \caption{Average 5-shot Accuracy on the \fewmany{} benchmark of various \fastfit{} models over training time, measured in seconds, trained on an Nvidia A100-80GB GPU.}
    \label{fig:accuracy_pre_step_5_shot}\vspace{-1em}
\end{figure}

\textbf{Convergence.}
Figure \ref{fig:accuracy_pre_step_5_shot} presents the average \fewmany{} accuracy results per training second of \fastfit{} for 5-shot with both small and large, and regular and ST backbone models. Results demonstrate \fastfit{} rapid convergence, achieving top performances within a few seconds before reaching a plateau. Notably, both small and large Sentence Transformer (ST) models exhibit higher initial performance and faster convergence than their non-ST base model counterparts. We can also see that \fastfit{} achieves state-of-the-art results on \fewmany{}, above 81.2, within 30 seconds as illustrated in Fig. \ref{fig:fastfit_summary}.



\section{Ablation \& Full Training}
To further examine the contribution of some of our method modifications, we compare training with CLS and token-level similarity metrics, as well as training with a different number of batch repetitions. We conduct these experiments on three datasets: Hwu64, Banking77, and Clinc150, with 5 random splits, and average their results. We assess the effect of these modifications for both small and large models, with 5 and 10 shots.


In Table \ref{tbl:abl}, we present the differences in performance caused by our changes; full results are available in App. \S\ref{sec:app:ablation_results}. The Token-level similarity metric proves beneficial across all settings, with a more pronounced effect for smaller models and when less data is available (5-shot compared to 10-shot). Concerning the number of repetitions, we observe that, in most cases, adding repetitions helps. Additionally, it appears that overall, four repetitions are more effective than two.
Regarding the relationship between the number of shots and the effectiveness of repetition, no clear connection is apparent. While an increase in the number of shots enhances effectiveness in small models, the opposite is observed for large models, where the effect decreases. Nevertheless, it seems that, in general, larger models benefit more from batch repetition.

Although our primary focus is few-shot classification, we also wanted to examine the effectiveness of \fastfit{} when training on the full dataset. We conducted two sets of experiments. In the first, we compared \fastfit{}-small, \fastfit{}-large, and a large standard classifier on Hwu64, Banking77, and Clinc150. In the second, we compared \fastfit{}-small and \fastfit{}-large with a few base-sized multilingual baseline models on MASSIVE, using the set of six languages mentioned in \S\ref{sec:multi-data}. These baselines are based on the MASSIVE paper, where Classifier-B and mT5-B Encoder are standard classifiers based on XLM-R-BASE and mT5-Base with 270M and 258M parameters, respectively. mT5-B T2T is a text-2-text classifier with 580M parameters.


Results in Table \ref{tbl:full} demonstrate that when training on all the data, \fastfit{}-Small outperforms the large Classifier, and \fastfit{}-Large performs even better. From Table \ref{tbl:full_multi}, we can see that \fastfit{}-Small outperforms all other baselines even with fewer than half the number of parameters. Moreover, \fastfit{}-Large further improves performances by $0.6\%$ on average. These results indicate that \fastfit{} is not only a fast few-shot classifier but can also outperform even larger classifiers when training on the full dataset.


\begin{table}[t!]
\centering
\resizebox{\columnwidth}{!}{
\begin{tabular}{@{}ccccc@{}}
\toprule
\multicolumn{1}{l}{Model} & \multicolumn{1}{l}{Shot} & \multicolumn{1}{l}{Similarity Level} & \multicolumn{2}{c}{Repetitions} \\ \midrule
  &    &   Token   & 2     & 4    \\ \midrule
\fastfit{}-S & 5  & 1.33 & -0.28 & 0.09 \\
\fastfit{}-S & 10 & 0.85 & 0.09  & 0.24 \\
\fastfit{}-L & 5  & 0.65 & 0.72  & 1.04 \\
\fastfit{}-L & 10 & 0.36 & 0.55  & 0.78 \\ \bottomrule
\end{tabular}}
\caption{\fastfit{} ablation experiments; Accuracy differences in training with token-level versus CLS similarity metrics and increasing augmentations repetitions. Token-level enhancements are more prominent in smaller models, especially in the 5-shot setting.}
\label{tbl:abl}\vspace{-0.5em}
\end{table}

\begin{table}[t!]
\centering
\begin{tabular}{@{}ccccc@{}}
\toprule
Model         & C150     & B77       & H64   & Avg. \\ \midrule
Classifier-L     & 96.8     & 93.7      & 92.1  & 94.2 \\
\fastfit{}-S & 97.1     & 93.8      & 92.7  & 94.5 \\
\fastfit{}-L & \textbf{97.5}     & \textbf{94.2}      & \textbf{93.0}    & \textbf{94.9} \\ \bottomrule
\end{tabular}
\caption{\fastfit{} accuracy results when training on the full data.}
\label{tbl:full}\vspace{-0.5em}
\end{table}

\begin{table}[t!]
\centering
\resizebox{\columnwidth}{!}{
\begin{tabular}{@{}cccccccc@{}}
\toprule
Model           & EN   & DE   & JP   & ES   & FR   & CN   & Avg. \\ \midrule
Classifier-B & 88.3 & 85.7 & 83.9 & 86.9 & 86.3 & 84.9 & 86.0 \\
mT5-B T2T       & 87.9 & 86.2 & 83.5 & 86.7 & 86.9 & 85.2 & 86.1 \\
mT5-B Enc       & 89.0   & 86.8 & 85.8 & 86.8 & 87.2 & 85.8 & 86.9 \\
\fastfit{}-S       & 88.8 & 87.4 & 87.0 & \textbf{87.9} & 87.6 & \textbf{86.7} & 87.6 \\
\fastfit{}-L       & \textbf{89.5} & \textbf{88.5} & \textbf{88.5} & 87.4 & \textbf{88.5} & \textbf{86.7} & \textbf{88.2} \\ \bottomrule
\end{tabular}}
\caption{\fastfit{} and baselines accuracy results on MASSIVE with full data training.}
\label{tbl:full_multi}\vspace{-0.5em}
\end{table}

\section{Related Work}
For fine-tuning baselines, we focus on readily available methods. , including SetFit with its package, a standard classifier accessible through HF Transformers \cite{Wolf2019HuggingFacesTS}, or LLMs through API calls.  However, there are various few-shot classifiers, and we will briefly discuss a couple of them. QAID \cite{yehudai2023qaid} proposed pre- and fine-tuning training stages with unsupervised and supervised loss, using ColBERT architecture, achieving SOTA results. T-Few \cite{Liu2022FewShotPF}, a parameter-efficient fine-tuning method based on T0 \cite{Sanh2021MultitaskPT}, claims to be better and cheaper than In-Context Learning.

Regarding few-shot prompting of LLMs approaches, a question arises about whether our results will withstand stronger LLMs or improved prompting techniques.  According to \citet{Loukas2023MakingLW} we can deduce that \fastfit{} outperforms GPT4 \cite{openai2023gpt4} with a fraction of the cost. Additionally, \citet{Milios2023InContextLF} demonstrate that retrieval-based few-shot prompts can lead to improved results. However, it's worth noting that currently, these models remain slow and costly.

\section{Conclusions}
In this paper, we introduce \fastfit{}, a novel few-shot text classification method accompanied by a Python package. For our task, we curated the \fewmany{} benchmark. Our results demonstrate that \fastfit{} outperforms large language models (LLMs) such as Flan-XXL and Llama-2-chat-70B, as well as fine-tuning methods, including both standard and SetFit classifiers, readily available in existing packages. Notably, \fastfit{} exhibits fast training and inference. We provide evidence that these results hold for both Multilingual and full-data training setups. We hope that \fastfit{}'s speed and simplicity will enhance its usability.

\bibliography{acl_latex}

\appendix
\newpage

\section{Full Code Example}
\label{sec:app:code_exmp}





    





Any dataset can be loaded directly from Huggingface Hub, For example:
\begin{mycodebox}
\begin{lstlisting}
from datasets import load_dataset

dataset = load_dataset(
    "FastFit/banking77"
)
\end{lstlisting}
\end{mycodebox}

Then \fastfit{} library can sample it down to the 5 or 10-shot format:
\begin{mycodebox}
\begin{lstlisting}
from fastfit import sample_dataset

dataset["train"] =sample_dataset(
    dataset["train"],
    label_column="label",
    num_samples_per_label=5
)
\end{lstlisting}
\end{mycodebox}

Then once the data is ready it can be serve as input to the Fast-Fit trainer together with other important inputs:

\begin{mycodebox}
\begin{lstlisting}
from fastfit import FastFitTrainer

trainer = FastFitTrainer(
    model_name_or_path=
    "roberta-large",
    label_column_name="label_text",
    text_column_name="text",   
    dataset=dataset, 
)

model = trainer.train()
results = trainer.evaluate()
\end{lstlisting}
\end{mycodebox}

Then we can save the model:
\begin{mycodebox}
\begin{lstlisting}
model.save_pretrained("fast-fit")
\end{lstlisting}
\end{mycodebox}


And could be loaded for inference with:

\begin{mycodebox}
\begin{lstlisting}
from fastfit import FastFit
from transformers import (
     AutoTokenizer, 
     pipeline
)

model = FastFit.from_pretrained(
    "fast-fit"
)

tokenizer =  \
    AutoTokenizer.from_pretrained(
        "roberta-large"
)

classifier = pipeline(
    "text-classification",
    model=model, 
    tokenizer=tokenizer
)

print(classifier("Hello World!"))
\end{lstlisting}
\end{mycodebox}


\begin{figure}[ht]
    \centering
    \resizebox{1.05\columnwidth}{!}{%
        \includegraphics{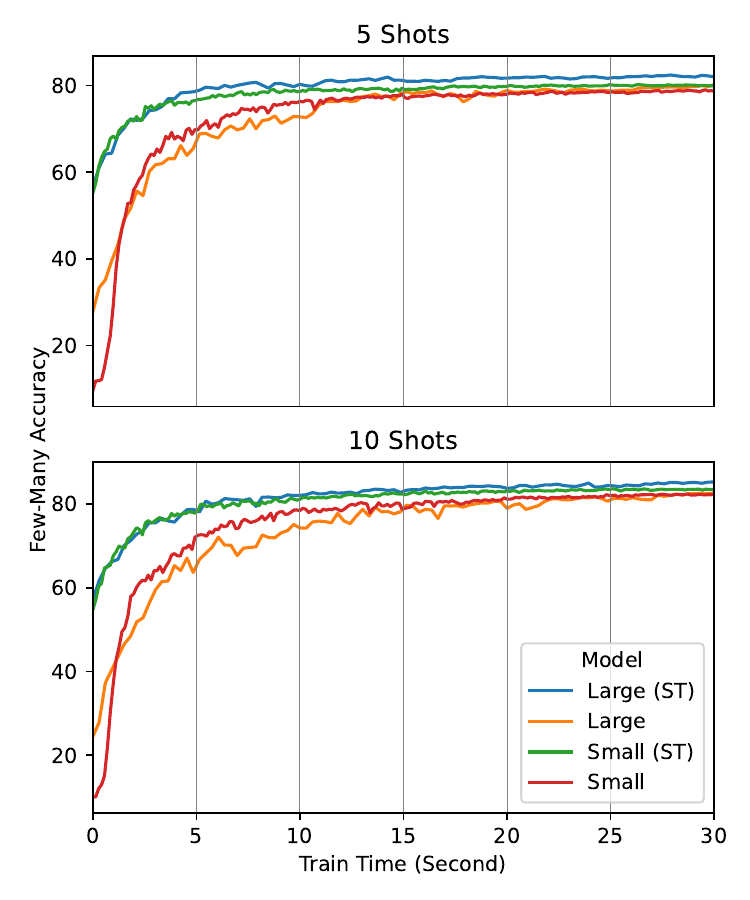}%
    }
    \caption{Average 5 and 10 shot Accuracy on the Few-Many benchmark of various \fastfit{} models over training time, measured in seconds, trained on an Nvidia A100-80GB GPU.}
    \label{fig:accuracy_pre_step_both}\vspace{-1em}
\end{figure}


\section{Data Statistics}\label{sec:app:data_stat}
We construct and experiment with \fewmany{} benchmark, containing 8 datasets with at least 50 classes, See Table \ref{tbl:data-stat}, for full data statistics.
Three English intent detection few-shot text classification datasets are: \textbf{Hwu64} \citep{liu2019benchmarking_3}, \textbf{Banking77} \citep{casanueva2020efficient_2}, and \textbf{Clinc150} \citep{larson2019evaluation_4}. Many classes are semantically similar, making the classification tasks much harder. Two datasets focus on topic classification in the debate context are: \textbf{Argument Topic} \cite{gretz2019largescale} and \textbf{Claim Stance} \cite{bar-haim-etal-2017-stance} containing arguments associated with debatable topics, where the debatable topic is the gold label.
\textbf{DBpedia} data set results from a project aiming to extract structured content from Wikipedia articles. Here we use the data set with L2 (L1-L3) level for text classification with 70 classes. 
The \textbf{Trec} dataset \cite{li-roth-2002-learning} presents the task of question classification, with 50 fine-grain classes describing the answer type.
\textbf{Amazon Products}(AP) is a classification data of product descriptions to 106 product categories.
 We conduct our experiments in 5/10-shot scenarios where in the $k$-shot scenario the training set consisted of $k$ examples per class.

\begin{table}[h]
\centering
\resizebox{\columnwidth}{!}{
\begin{tabular}{lccccc}
\toprule 
Dataset & \#Train & \#Vaild & \#Test & \#Classes & \#Domains\\
\midrule
Clinc150    & 15,000  & 3,000 & 4,500 & 150 & 10\\
BankingG77  & 8,622   & 1,540 & 3,080 & 77 & 1\\
Hwu64       & 8,954   & 1,076 & 1,076 & 64 & 21\\
Argument Topic    & 6,640  & 949 & 1,898 & 71 & -\\
Claim Stance  & 1,675   & - & 480 & 55 & -\\
DBpedia       & 240,942   & 36,003 & 60,794 & 70 & -\\
Amazon Products     & 22,036  & 2,459 &  6,148 & 106 & -\\
Trec     & 5,452  & - &  500 & 50 & -\\
MASSIVE     & 11,514  & 2,033 &  2,974 & 60 & 18\\
\bottomrule
\end{tabular}}
\caption{
Data statistics of the few-shot classification datasets.
}
\label{tbl:data-stat}
\end{table}


\section{Full \fewmany{} Results}\label{sec:app:full_fewmany}
\textbf{LLMs in-context examples.} All LLMs, except Mistral, have a context window of 4K tokens. For Clinc150, Banking77, and Argument Topic, we fit 1 example into their context; for Trec and Claim Stance, we fit 2 examples; and for Hwu64, we fit 3 examples. Mistral, with an 8K context window, allows for 2 examples for Clinc150, 3 for Banking77 and Argument Topic, and 5 examples for all the remaining test sets.

Table \ref{tbl:fewmany-abl} presents the comprehensive results of \fewmany{} across its 8 diverse test sets, covering 5- and 10-shot settings, with small (MPNet) and large (RoBERTa) models, as well as regular and Sentence Transformers (ST) backbone models. The results demonstrate that \fastfit{} consistently outperforms SetFit and standard classifiers on average across all settings. In the 5-shot setting, \fastfit{} achieves $2\%$ and $4.5\%$ higher scores than SetFit and standard classifiers, respectively. Similarly, in the 10-shot case, it surpasses them by $2.7\%$ and $0.7\%$. Furthermore, we observe that large and ST models consistently outperform their small and regular counterparts.

Table \ref{tbl:st-improvment-abl-small} shows the performance differences between models with and without ST for 5- and 10-shots, using both small and large models. The results are averaged across all \fewmany{} test sets. We observe that the difference is consistently more significant for 5-shot compared to 10-shot, indicating that when fewer examples are available, the backbone model becomes more advantageous. Moreover, the difference is more pronounced for the large models in \fastfit{} and SetFit, suggesting that large ST models enable even greater improvement. Finally, we note that the differences are smaller for \fastfit{} compared to SetFit, implying that \fastfit{} is less reliant on ST backbone models than SetFit. These findings are consistent with our results from the multilingual experiment and highlight the adaptability of \fastfit{}.

\section{Multilingual Results}\label{sec:app:multi_results}
In Table \ref{tbl:multi-abl-setfit}, we present the experimental results using various backbone models for SetFit. We evaluated three models: (1) Monolingual sentence-transformer (ST) large, referred to as ST-L. (2) Regular Multilingual RoBERTa-large, denoted as XLM-R-L or simply L. (3) RoBERTa-Base Multilingual sentence-transformer model, labeled as ST-XB.

The results indicate that ST-L encounters difficulties with all non-English datasets, resulting in overall inferior performance. XLM-R-L exhibits lower proficiency in English but demonstrates improved results across all other languages. Lastly, ST-XB, with a comparable size to the small models (125M vs. 110M), achieved similar, albeit slightly lower, results. These findings underscore SetFit's dependence on ST pre-trained models and highlight its limitations when such a model is unavailable, as in this experiment.

\section{Training Run Times Results}\label{sec:app:runtime}
\begin{figure}[ht]
    \centering
    \resizebox{0.85\columnwidth}{!}{%
        \includegraphics{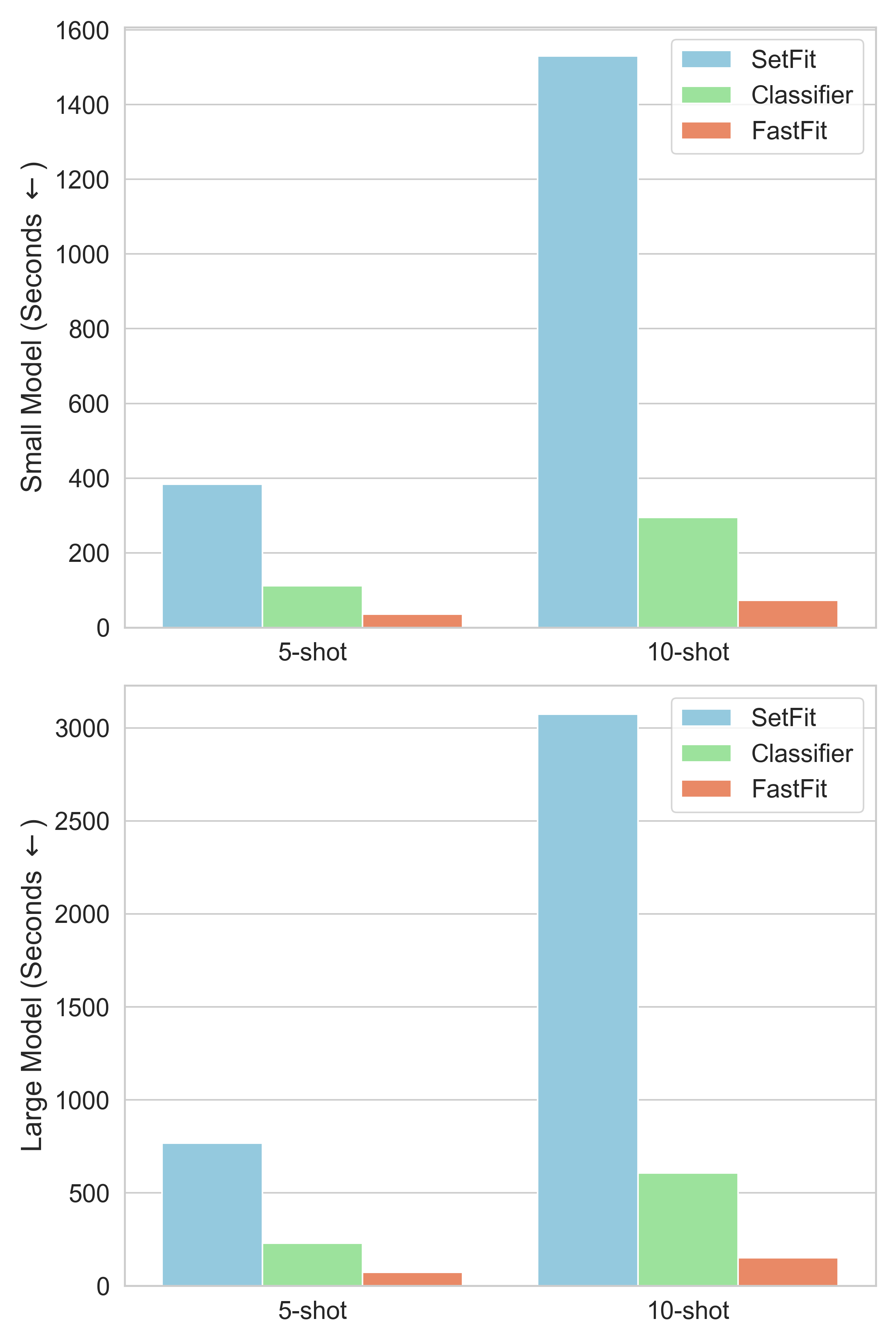}%
    }
    \caption{Training times (sec) for \fastfit{}, SetFit, and standard classifier. \fastfit{} training is faster both for the small model (top) and for the large model (bottom).}
    \label{fig:fastfit_extra_time}\vspace{-1em}
\end{figure}

Here we present more training run time results for \fastfit{}, SetFit, and a standard classifier. Fig. \ref{fig:fastfit_extra_time} presents the run time for the small and large settings. Tab. \ref{tbl:runtime} shows the average training run time results. Table \ref{fig:accuracy_pre_step_both} presents the convergence in both 5- and 10-shot cases. Table \ref{tbl:app:full_train_time} further shows the training times for the different methods, models, and datasets.


\begin{table}[ht]
    \centering
\resizebox{\columnwidth}{!}{%
    \begin{tabular}{lcccc}
        \toprule
        Model & \multicolumn{2}{c}{Small} & \multicolumn{2}{c}{Large} \\
        \cmidrule(lr){2-3} \cmidrule(lr){4-5}
        & 5-shot & 10-shot & 5-shot & 10-shot \\
        \midrule
        \fastfit{} & 00:01:16$^*$ & 00:02:11$^*$ & 00:02:44$^*$ & 00:05:20$^*$  \\
        SetFit & 00:12:44 & 00:49:38 & 00:25:36 & 01:34:18 \\
        Classifier & 00:02:20 & 00:04:31 & 00:05:44 & 00:11:04 \\
        \bottomrule
    \end{tabular}
    }
            \caption{Training Times for \fastfit{}, SetFit, and Standard Classifier: \fastfit{} times are indicated with a $*$ to denote that, as illustrated in Figure \ref{fig:accuracy_pre_step_5_shot}, they converge after approximately 30 seconds and promptly reach a plateau thereafter. Comprehensive training times for all models and datasets are presented in Table \ref{tbl:app:full_train_time}.}\label{tbl:runtime}
\end{table}

\newpage
\section{Ablation Results}\label{sec:app:ablation_results}
Here, we present the results for the ablations associated with Table \ref{tbl:abl}. The first ablation is designed to measure the effect of the similarity metrics. Table \ref{tbl:abl-sim} shows the results of the experiments with both CLS and token-level similarity metrics. In Table \ref{tbl:abl-rep}, we present the results without augmentation repetitions (1), and with 2 and 4 repetitions. Both ablations support our claim that the token-level similarity metric and an increased number of augmentation repetitions help.

\section{Short Video}\label{sec:app:vid}
Click \href{https://www.youtube.com/watch?v=UqLGxpnd5YA}{here} for our short presentation.

\begin{table*}[t!]
\centering

\begin{tabular}{@{}p{2cm}p{1.5cm}p{1.2cm}p{1.2cm}p{1.2cm}p{1.2cm}p{1.2cm}p{1.2cm}p{1.5cm}@{}}
\toprule
\multicolumn{1}{l}{Method} & Model  & En   & De   & Ja   & Es   & Fr   & Zh   & Average \\ \midrule
                           &       & \multicolumn{6}{c}{5-shot}              &         \\ \midrule
\multirow{2}{*}{\fastfit{}}& S & \underline{72.3} & \underline{65.0} & \underline{68.7} & \underline{65.9} & \underline{68.0} & \underline{68.4} & \underline{68.1} \\
                        & L & \textbf{77.6*} & \textbf{70.5*} & \textbf{73.7*} & \textbf{71.7*} & \textbf{73.1*} & \textbf{73.7*} & \textbf{73.4*}  \\
SetFit & S & 67.9 & 62.2 & 66.8 & 64.0 & 65.0 & 66.7 & 65.4   \\
                        & ST-L  & 74.0 & 50.3 & 41.3 & 53.6 & 52.1 & 39.6 & 51.8    \\
                        & L & 66.1 & 60.8 & 64.8 & 50.1 & 61.3 & 43.6 & 57.8    \\
                        & ST-XB & 74.0 & 62.3 & 64.8 & 62.0 & 62.3 & 65.1 & 65.1    \\
                        \midrule
                        &       & \multicolumn{6}{c}{10-shot}             &         \\ \midrule
\multirow{2}{*}{\fastfit{}}& S & \underline{77.6} & 70.5 & 73.7 & \underline{71.7} & \underline{73.1} & \underline{73.7} & \underline{73.4}    \\
                        & L & \textbf{79.2*} & \textbf{74.8*} & \textbf{77.4} & \textbf{74.1*} & \textbf{75.7*} & \textbf{74.9*} & \textbf{76.0*}    \\
SetFit & S & 74.7 & 69.8 & 73.5 & 71.4 & 72.0 & 72.9 & 72.4    \\
                        & ST-L  & 78.3 & 61.4 & 53.4 & 64.0 & 63.2 & 48.3 & 61.4    \\
                        & L & 74.5 & 69.1 & 72.5 & 69.7 & 70.7 & 59.2 & 69.3    \\
                        & ST-XB & 78.3 & 68.7 & 72.9 & 70.1 & 70.5 & 72.3 & 72.1    \\
                        \bottomrule
\end{tabular}
\caption{Accuracy results for \fastfit{} and baselines across six languages, under 5/10-shot settings. Results with few SetFit versions but no one surpasses SetFit small. We experimenting here with sentence-transformer (ST) large monolingual, multilingual base, and non-ST multilingual large.
}\vspace{-1em}
\label{tbl:multi-abl-setfit}
\end{table*}

\begin{table*}[t!]
\centering
\begin{tabular}{@{}p{1.6cm}p{1.4cm}p{1.0cm}p{1.0cm}p{1.0cm}p{1.0cm}p{1.0cm}p{1.0cm}p{1.0cm}p{1.0cm}p{1.2cm}@{}}
\toprule
\multicolumn{1}{l}{Method} & Model & C150 & AP106 & B77 & AT71 & D70 & HU64 & CS55 & T50 & Average \\ \midrule
& & \multicolumn{8}{c}{5-shot} & \\ \midrule
\multirow{4}{*}{\fastfit{}}& S & 89.4 & 46.7 & 80.2 & 95.7 & 81.3 & 81.3 & 85.8 & 73.9 & 79.3 \\
& S (ST) & 91.3 & 47.5 & 81.0 & 95.4 & 82.5 & 82.2 & 86.1 & 80.3 & 80.8 \\
& L & 91.9 & 50.0 & 83.1 & 96.1 & 81.5 & 82.6 & 85.9 & 80.5 & 81.4 \\
& L (ST) & \textbf{93.4} & \textbf{50.9} & \textbf{85.2} & \textbf{96.2} & \textbf{83.1} & \textbf{84.6} & 87.5 & \textbf{84.8} & \textbf{83.2} \\
\midrule
\multirow{4}{*}{SetFit} & S & 87.8 & 43.0 & 75.7 & 94.3 & 79.2 & 76.3 & 84.0 & 74.9 & 76.9 \\
& S (ST) & 89.0 & 45.9 & 77.3 & 94.8 & 79.0 & 80.0 & 84.1 & 79.5 & 78.7 \\
& L & 84.5 & 45.9 & 79.2 & 94.7 & 80.1 & 79.5 & 84.5 & 78.3 & 78.3 \\
& L (ST) & 90.4 & 48.2 & 81.7 & 95.6 & 80.1 & 81.9 & 87.8 & 83.9 & 81.2 \\
\midrule
\multirow{4}{*}{Classifier} & S & 75.9 & 25.7 & 58.1 & 94.3 & 67.6 & 61.4 & 73.5 & 54.2 & 63.8 \\
& S (ST) & 86.3 & 30.4 & 68.2 & 95.1 & 70.5 & 73.9 & 82.6 & 63.4 & 71.3\\
& L & 89.9 & 46.0 & 74.9 & 95.6 & 78.4 & 77.2 & 86.1 & 66.9 & 76.9 \\
& L (ST) & 92.0 & 44.5 & 79.7 & 96.0 & 76.8 & 79.4 & \textbf{88.2} & 73.3 & 78.7 \\

                \midrule
                &       & \multicolumn{8}{c}{10-shot} & \\ \midrule
\multirow{4}{*}{\fastfit{}}& S & 93.3 & 53.9 & 85.9 & 96.3 & 86.6 & 86.0 & 87.9 & 82.9 & 84.1 \\
& S (ST) & 93.5 & 54.5 & 86.4 & 95.9 & 87.8 & 85.8 & 88.5 & 84.1 & 84.6 \\
& L & 94.1 & 56.8 & 87.8 & 96.4 & 87.0 & 86.2 & 88.2 & 86.3 & 85.4 \\
& L (ST) & \textbf{95.3} & \textbf{57.5} & \textbf{88.8} & 96.5 & \textbf{88.7} & \textbf{87.9} & 89.4 & \textbf{88.0} & \textbf{86.5} \\
\midrule
\multirow{4}{*}{SetFit} & S & 90.0 & 53.1 & 84.4 & 95.2 & 84.9 & 84.0 & 87.4 & 83.0 & 82.8 \\
& S (ST) & 90.9 & 53.6 & 84.8 & 95.5 & 85.9 & 85.1 & 87.7 & 83.7 & 83.4 \\
& L & 78.5 & 52.5 & 84.4 & 94.3 & 85.0 & 83.2 & 86.1 & 84.6 & 81.1 \\
& L (ST) & 88.4 & 53.6 & 86.4 & 95.7 & 85.8 & 85.4 & 88.8 & 86.4 & 83.8 \\
\midrule
\multirow{4}{*}{Classifier} & S & 88.1 & 43.6 & 75.6 & 95.3 & 80.1 & 75.9 & 84.1 & 68.0 & 76.3 \\
& S (ST) & 91.5 & 46.9 & 80.2 & 95.5 & 82.1 & 83.1 & 86.5 & 78.0 & 80.5 \\
& L & 93.5 & 57.7 & 86.1 & \textbf{96.6} & 87.3 & 85.4 & 88.9 & 83.1 & 84.8 \\
& L (ST) & 94.5 & 57.1 & 87.4 & \textbf{96.6} & 87.0 & 86.0 & \textbf{90.9} & 86.8 & 85.8 \\
\bottomrule
\end{tabular}
\caption{Accuracy results for \fastfit{} and baselines across \fewmany{} benchmark tasks, under 5/10-shot settings. Small model (S) is MPNet and (L) is RoBERTa Large, Sentence Transformers models are marked with (ST).
}\vspace{-1em}
\label{tbl:fewmany-abl}
\end{table*}

\begin{table*}[t!]
\centering
\resizebox{\columnwidth}{!}{%
\begin{tabular}{@{}lcccccc@{}}
\toprule
Method & \multicolumn{3}{c}{5-shot} & \multicolumn{3}{c}{10-shot} \\
 &  Base    & ST   & Diff  & Base   & ST   & Diff  \\
\midrule
 & \multicolumn{6}{c}{Small Model} \\
\midrule
\multirow{1}{*}{\fastfit{}} & 79.3 & 80.8 & +1.5 & 84.1 & 84.6 & +0.5\\
\multirow{1}{*}{SetFit} &  76.9 & 78.7 & +1.8 & 82.8 & 83.4 & +0.6 \\
\multirow{1}{*}{Classifier} & 63.8 & 71.3 & +7.5 & 76.3 & 80.5 & +4.2 \\

\midrule
 & \multicolumn{6}{c}{Large Model} \\
\midrule
\multirow{1}{*}{\fastfit{}} & 81.4 & 83.2 & +1.8 & 85.4 & 86.5 & +1.1\\
\multirow{1}{*}{SetFit} &  78.3 & 81.2 & +2.9 & 81.1 & 83.8 & +2.7 \\
\multirow{1}{*}{Classifier} & 76.9 & 78.7 & +1.8 & 84.8 & 85.8 & +1.0 \\
\bottomrule

\end{tabular}
}
\caption{Accuracy results with Sentence Transformers (ST) regular backbone model for \fastfit{} and baselines on \fewmany{}, under 5/10-shot and small/large model.  The difference (Diff) column represents the improvement due to the use of ST backbone model. Model S is MPNet and L is RoBERTa Large.
}
\label{tbl:st-improvment-abl-small}
\end{table*}

\begin{table*}[t!]
\centering
\begin{tabular}{@{}p{2cm}p{1.5cm}p{1.2cm}p{1.2cm}p{1.2cm}p{1.2cm}p{1.2cm}@{}}
\toprule
Method & Shots & Sim. metric & C150 & B77 & H64 & Average \\
\midrule
\multirow{4}{*}{\fastfit{}-small} & 5 & CLS & 88.9 & 78.6 & 78.5 & 82.0 \\
                              & 5 & TOK. & 90.2 & 80.0 & 79.7 & \textbf{83.3} \\
                              & 10 & CLS & 92.4 & 84.7 & 83.8 & 86.9 \\
                              & 10 & TOK. & 93.3 & 85.4 & 84.7 & \textbf{87.8} \\
\multirow{4}{*}{\fastfit{}-large} & 5 & CLS & 91.6 & 81.7 & 82.4 & 85.2 \\
                               & 5 & TOK. & 92.3 & 82.9 & 82.4 & \textbf{85.9} \\
                               & 10 & CLS & 94.1 & 87.6 & 86.3 & 89.4 \\
                               & 10 & TOK. & 94.8 & 88.0 & 86.4 & \textbf{89.7} \\
\bottomrule
\end{tabular}
\caption{Ablation results with CLS and token-level similarity metrics. The average results that scored the highest for each model size and shot number are highlighted in bold.}
\label{tbl:abl-sim}
\end{table*}

\begin{table*}[t!]
\centering
\begin{tabular}{@{}p{2cm}p{1.5cm}p{1.5cm}p{1.2cm}p{1.2cm}p{1.2cm}p{1.2cm}p{1.2cm}@{}}
\toprule
Method & Size & Shots & Repet. & C150 & B77 & H64 & Average \\
\midrule
\multirow{3}{*}{\fastfit{}} & \multirow{3}{*}{S} & \multirow{3}{*}{5} & 1 & 90.3 & 80.3 & 79.1 & 83.2 \\
                              && & 2 & 89.8 & 79.8 & 79.2 & 82.9 \\
                              && & 4 & 90.2 & 80.0 & 79.7 & \textbf{83.3} \\
                              \midrule
\multirow{3}{*}{\fastfit{}} & \multirow{3}{*}{S} & \multirow{3}{*}{10} & 1 & 93.3 & 85.3 & 84.1 & 87.6 \\
                          &    &  & 2 & 93.2 & 85.3 & 84.5 & 87.6 \\
                           &   &  & 4 & 93.3 & 85.4 & 84.7 & \textbf{87.8} \\
                           \midrule
\multirow{3}{*}{\fastfit{}} & \multirow{3}{*}{L} & \multirow{3}{*}{5} & 1 & 91.6 & 82.0 & 81.0 & 84.8 \\
                             & &  & 2 & 92.0 & 82.4 & 82.3 & 85.6 \\
                             & &  & 4 & 92.3 & 82.9 & 82.4 & \textbf{85.9} \\
                             \midrule
\multirow{3}{*}{\fastfit{}} & \multirow{3}{*}{L} & \multirow{3}{*}{10} & 1 & 94.2 & 87.3 & 85.2 & 88.9 \\
                           &   &  & 2 & 94.6 & 87.7 & 86.1 & 89.5 \\
                            &  &  & 4 & 94.8 & 88.0 & 86.4 & \textbf{89.7} \\
\bottomrule
\end{tabular}
\caption{Ablation results with varying repetition numbers. The bolded values represent the highest-scoring average results for each model size and shot number.}
\label{tbl:abl-rep}
\end{table*}

\begin{table*}[t!]
\centering
\resizebox{\textwidth}{!}{%
\begin{tabular}{lrlllllllll}
\toprule
Model & Shots & C150 & AP106 & B77 & AT71 & D70 & CS55 & HU64 & T50 & Average \\
\midrule
\fastfit{} Small (ST) (ML) & 10 & 00:03:41 & 00:03:55 & 00:02:59 & 00:02:13 & 00:03:20 & 00:01:35 & 00:01:29 & 00:01:15 & 00:02:34 \\
\fastfit{} Small (ST) (ML) & 5 & 00:01:49 & 00:02:22 & 00:01:29 & 00:01:08 & 00:01:40 & 00:00:49 & 00:00:44 & 00:00:39 & 00:01:20 \\
\fastfit{} Small (ST) & 10 & 00:03:26 & 00:03:50 & 00:02:55 & 00:02:07 & 00:03:26 & 00:01:23 & 00:01:25 & 00:01:10 & 00:02:28 \\
\fastfit{} Small (ST) & 5 & 00:01:43 & 00:02:11 & 00:01:27 & 00:01:04 & 00:01:43 & 00:00:43 & 00:00:42 & 00:00:35 & 00:01:16 \\
\fastfit{} Small & 10 & 00:03:28 & 00:03:49 & 00:02:56 & 00:02:05 & 00:03:26 & 00:01:24 & 00:01:23 & 00:01:09 & 00:02:28 \\
\fastfit{} Small & 5 & 00:01:42 & 00:02:12 & 00:01:27 & 00:01:04 & 00:01:42 & 00:00:43 & 00:00:41 & 00:00:35 & 00:01:16 \\
\fastfit{} Large (ST) & 10 & 00:07:30 & 00:08:36 & 00:06:18 & 00:04:18 & 00:07:40 & 00:03:06 & 00:02:50 & 00:02:24 & 00:05:20 \\
\fastfit{} Large (ST) & 5 & 00:03:36 & 00:04:58 & 00:03:07 & 00:02:12 & 00:03:50 & 00:01:35 & 00:01:24 & 00:01:13 & 00:02:44 \\
\fastfit{} Large (ML) & 10 & 00:07:53 & 00:09:06 & 00:06:39 & 00:04:51 & 00:07:48 & 00:03:29 & 00:03:08 & 00:02:46 & 00:05:42 \\
\fastfit{} Large (ML) & 5 & 00:03:49 & 00:05:26 & 00:03:18 & 00:02:27 & 00:03:54 & 00:01:47 & 00:01:31 & 00:01:23 & 00:02:57 \\
\fastfit{} Large & 10 & 00:07:26 & 00:08:39 & 00:06:17 & 00:04:20 & 00:07:39 & 00:03:04 & 00:02:48 & 00:02:25 & 00:05:20 \\
\fastfit{} Large & 5 & 00:03:35 & 00:04:58 & 00:03:07 & 00:02:11 & 00:03:50 & 00:01:34 & 00:01:23 & 00:01:14 & 00:02:44 \\
\bottomrule
\toprule
Model & Shots & C150 & AP106 & B77 & AT71 & D70 & CS55 & HU64 & T50 & Average \\
\midrule
SetFit Small (ST) (ML) & 10 & 02:28:32 & 00:42:54 & 00:40:07 & 00:34:20 & 00:40:28 & 00:17:50 & 00:26:42 & 00:14:44 & 00:45:42 \\
SetFit Small (ST) (ML) & 5 & 00:36:46 & 00:15:04 & 00:10:08 & 00:08:30 & 00:10:02 & 00:04:49 & 00:06:43 & 00:04:03 & 00:12:01 \\
SetFit Small (ST) & 10 & 02:14:22 & 00:38:36 & 00:34:47 & 00:31:11 & 01:45:01 & 00:15:10 & 00:24:44 & 00:13:10 & 00:49:38 \\
SetFit Small (ST) & 5 & 00:32:48 & 00:13:39 & 00:08:48 & 00:07:51 & 00:25:12 & 00:04:13 & 00:05:51 & 00:03:34 & 00:12:44 \\
SetFit Small & 10 & 02:10:12 & 00:37:33 & 00:35:45 & 00:30:50 & 01:42:14 & 00:15:45 & 00:23:20 & 00:12:56 & 00:48:34 \\
SetFit Small & 5 & 00:32:46 & 00:13:09 & 00:08:56 & 00:07:47 & 00:25:24 & 00:04:11 & 00:05:43 & 00:03:29 & 00:12:41 \\
SetFit Large (ST) & 10 & 04:17:39 & 01:25:54 & 01:13:47 & 01:02:28 & 02:46:13 & 00:33:44 & 00:47:50 & 00:26:48 & 01:34:18 \\
SetFit Large (ST) & 5 & 01:10:34 & 00:30:01 & 00:18:27 & 00:15:39 & 00:41:24 & 00:08:51 & 00:12:19 & 00:07:37 & 00:25:36 \\
SetFit Large (ML) & 10 & 04:50:47 & 01:36:26 & 01:24:32 & 01:11:00 & 04:51:33 & 00:36:50 & 00:52:49 & 00:30:11 & 01:59:16 \\
SetFit Large (ML) & 5 & 01:12:07 & 00:33:54 & 00:20:46 & 00:17:41 & 01:09:16 & 00:09:59 & 00:13:08 & 00:07:59 & 00:30:36 \\
SetFit Large & 10 & 04:18:11 & 01:25:50 & 01:13:44 & 01:02:42 & 04:25:02 & 00:32:15 & 00:47:25 & 00:26:36 & 01:46:28 \\
SetFit Large & 5 & 01:04:12 & 00:30:21 & 00:18:42 & 00:15:45 & 01:05:24 & 00:08:42 & 00:11:50 & 00:07:18 & 00:27:47 \\
\bottomrule
\toprule
Model & Shots & C150 & AP106 & B77 & AT71 & D70 & CS55 & HU64 & T50 & Average \\
\midrule
Classifier Small (ST) (ML) & 10 & 00:08:39 & 00:04:20 & 00:04:28 & 00:04:06 & 00:04:03 & 00:03:01 & 00:03:41 & 00:02:45 & 00:04:23 \\
Classifier Small (ST) (ML) & 5 & 00:04:19 & 00:02:34 & 00:02:14 & 00:02:03 & 00:02:01 & 00:01:34 & 00:01:50 & 00:01:26 & 00:02:15 \\
Classifier Small (ST) & 10 & 00:08:54 & 00:04:28 & 00:04:36 & 00:04:14 & 00:04:09 & 00:03:06 & 00:03:48 & 00:02:51 & 00:04:31 \\
Classifier Small (ST) & 5 & 00:04:27 & 00:02:39 & 00:02:18 & 00:02:07 & 00:02:05 & 00:01:37 & 00:01:54 & 00:01:30 & 00:02:20 \\
Classifier Small & 10 & 00:08:56 & 00:04:27 & 00:04:34 & 00:04:13 & 00:04:09 & 00:03:05 & 00:03:47 & 00:02:51 & 00:04:30 \\
Classifier Small & 5 & 00:04:27 & 00:02:38 & 00:02:18 & 00:02:07 & 00:02:05 & 00:01:37 & 00:01:54 & 00:01:29 & 00:02:19 \\
Classifier Large (ST) & 10 & 00:21:52 & 00:11:00 & 00:11:14 & 00:10:20 & 00:10:14 & 00:07:35 & 00:09:19 & 00:06:59 & 00:11:04 \\
Classifier Large (ST) & 5 & 00:10:58 & 00:06:33 & 00:05:40 & 00:05:12 & 00:05:08 & 00:03:57 & 00:04:41 & 00:03:39 & 00:05:44 \\
Classifier Large (ML) & 10 & 00:22:44 & 00:11:23 & 00:11:43 & 00:10:44 & 00:10:35 & 00:07:54 & 00:09:42 & 00:07:16 & 00:11:30 \\
Classifier Large (ML) & 5 & 00:11:23 & 00:06:46 & 00:05:54 & 00:05:26 & 00:05:19 & 00:04:07 & 00:04:51 & 00:03:47 & 00:05:57 \\
Classifier Large & 10 & 00:21:52 & 00:11:00 & 00:11:17 & 00:10:21 & 00:10:15 & 00:07:38 & 00:09:20 & 00:06:59 & 00:11:05 \\
Classifier Large & 5 & 00:10:59 & 00:06:33 & 00:05:41 & 00:05:13 & 00:05:09 & 00:03:58 & 00:04:42 & 00:03:40 & 00:05:44 \\
\bottomrule
\end{tabular}
}
\caption{Average Training time over 10 seeds for the different methods.}
\label{tbl:app:full_train_time}
\end{table*}

\end{document}